%% file: main.tex
\documentclass{article}

\usepackage{microtype}
\usepackage{graphicx}
\usepackage{subfigure}
\usepackage{booktabs}
\usepackage{enumitem}
\usepackage{wrapfig}
\usepackage{caption}

\usepackage{hyperref}
\usepackage[table,xcdraw,dvipsnames]{xcolor}

\usepackage[accepted]{icml2024}

\usepackage{amsmath}
\usepackage{amssymb}
\usepackage{mathtools}
\usepackage{amsthm}

\usepackage{pgfplots}
\usepackage{pgfplotstable}
\pgfplotsset{compat=1.18}

\usepackage[capitalize,noabbrev]{cleveref}

\theoremstyle{plain}

\theoremstyle{definition}

\theoremstyle{remark}

\usepackage[textsize=tiny]{todonotes}

\usepackage{multirow}

\begin{document}
\twocolumn[
\icmltitle{BSA: Ball Sparse Attention for Large-scale Geometries}

\icmlsetsymbol{equal}{*}

\begin{icmlauthorlist}
\icmlauthor{Catalin E. Brita}{equal,yyy}
\icmlauthor{Hieu Nguyen}{equal,yyy}
\icmlauthor{Lohithsai Yadala Chanchu}{equal,yyy}
\icmlauthor{Domonkos Nagy}{equal,yyy}
\icmlauthor{Maksim Zhdanov}{amlab}

\end{icmlauthorlist}

\icmlaffiliation{yyy}{Informatics Institute, University of Amsterdam}
\icmlaffiliation{amlab}{AMLab, University of Amsterdam}

\icmlcorrespondingauthor{}{\{catalin.brita, hieu.nguyen, lohithsai.yadala.chanchu, domonkos.nagy2\}@student.uva.nl, m.zhdanov@uva.nl}

\icmlkeywords{Machine Learning, ICML}

\vskip 0.3in
]

\printAffiliationsAndNotice{\icmlEqualContribution}

\begin{abstract}
\input{sections/0_abstract}
\end{abstract}
\begin{figure}
  \centering
  \includegraphics[width=0.48\textwidth]{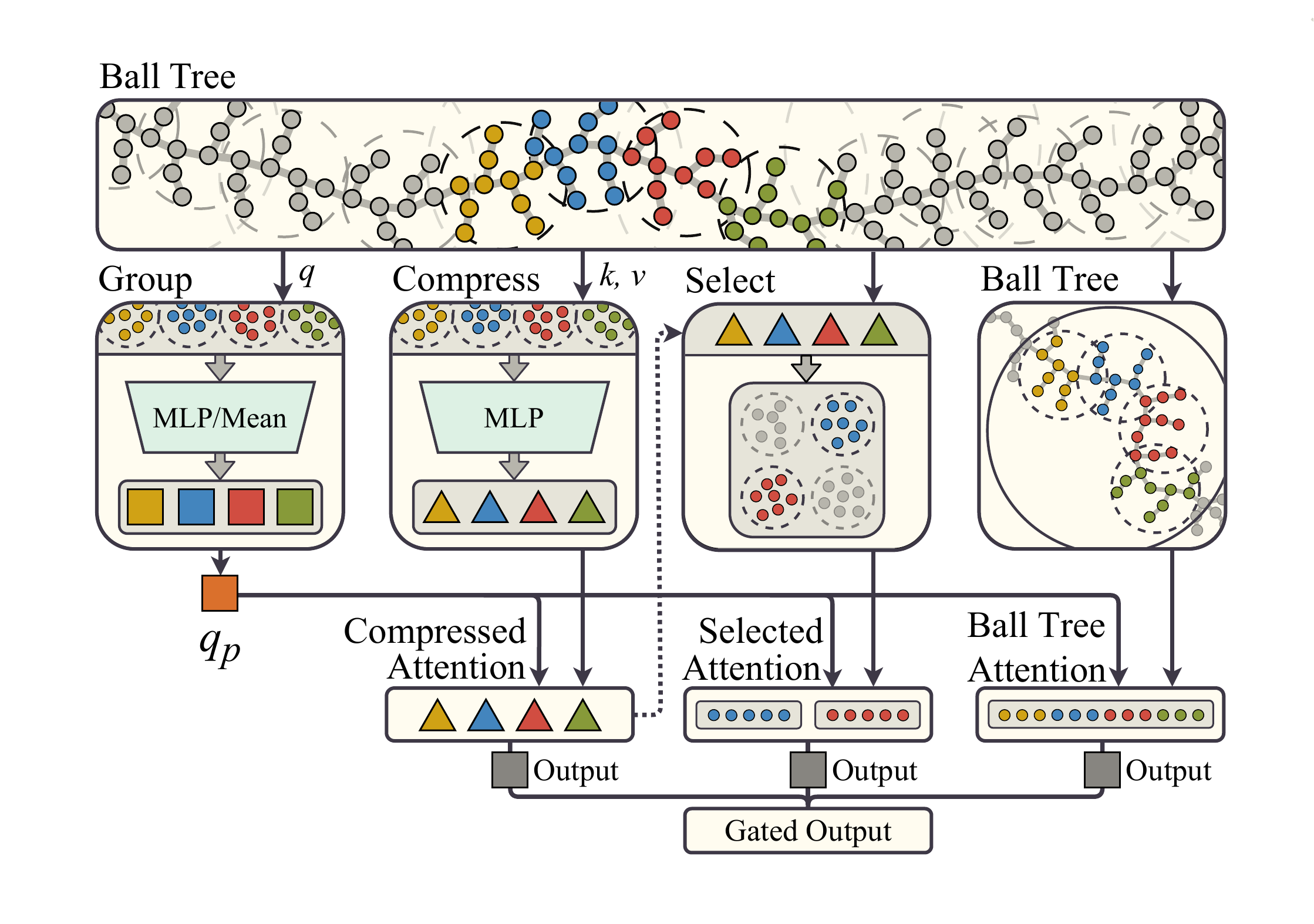}
  \caption{Ball Sparse Attention (BSA) pipeline. A ball tree imposes spatial locality, then three sparse‐attention branches: grouping (clustering points for shared selection), compression (MLP‐based token pooling), and selection (top‐k block retrieval) operate alongside fine‐grained Ball Tree Attention. A learnable gate fuses their outputs into the final attention.}
  \label{fig:figure1}
\end{figure}

\input{sections/1_introduction}
\input{sections/2_methodology}
\input{sections/3_results}
\input{sections/4_conclusion}

\bibliography{ref}
\bibliographystyle{icml2024}

\newpage
\appendix
\input{sections/5_appendix}

\end{document}

%% file: sections/0_abstract.tex
Self-attention scales quadratically with input size, limiting its use for large-scale physical systems. Although sparse attention mechanisms provide a viable alternative, they are primarily designed for regular structures such as text or images, making them inapplicable for irregular geometries. In this work, we present Ball Sparse Attention (BSA), which adapts Native Sparse Attention (NSA) \cite{yuan2025nativesparseattentionhardwarealigned} to unordered point sets by imposing regularity using the Ball Tree structure from the Erwin Transformer~\cite{zhdanov2025erwintreebasedhierarchicaltransformer}. We modify NSA's components to work with ball-based neighborhoods, yielding a global receptive field at sub-quadratic cost. On an airflow pressure prediction task, we achieve accuracy comparable to Full Attention while significantly reducing the theoretical computational complexity. We open-source our implementation~\footnote{\url{https://github.com/britacatalin/bsa}}.

%% file: sections/1_introduction.tex
\section{Introduction}
Scientific applications such as climate modeling \cite{curran2024resolutionagnostic}, molecule property prediction \cite{pengmei2023transformersefficienthierarchicalchemical}, or fluid flow simulation \cite{pengfluid}
increasingly rely on transformer-based architectures to capture complex, long-range dependencies in irregular data \cite{Abramson2024, Chen2025, Luo2024, Miao2024}. However, standard self-attention scales quadratically with input size, making it impractical for large-scale tasks in the scientific domain. This has motivated the development of scalable strategies for large-scale physical systems.

Sparse attention mechanisms mitigate quadratic scaling by computing attention for a strategically chosen subset of token pairs. These range from predefined or random sparsity patterns (e.g., BigBird \cite{zaheer2021bigbirdtransformerslonger}, where selection is random) to learned, data-dependent sparsity, as seen in Native Sparse Attention (NSA) \cite{yuan2025nativesparseattentionhardwarealigned}. NSA can select and compress tokens across the full sequence, allowing it to capture fine global dependencies. 

However, NSA is designed to work with text sequences that have a regular structure, while physical systems are often defined on irregular geometries. This poses a challenge as such structures are represented as unordered sets that do not have a canonical ordering that sparse methods utilize. Many approaches \cite{flatformer, swformer, pointtransformer} induce regular structure and transform point clouds into sequences to serve as inputs for sparse attention. One such approach is Point Transformer V3 \cite{pointtransformer}, which serializes points along a space-filling curve and partitions the resulting sequence into chunks for attention computation. Another solution is Octformer \cite{octformer}, which uses octrees to divide point clouds into local windows, each containing a fixed number of points.

Recently proposed Erwin \cite{zhdanov2025erwintreebasedhierarchicaltransformer} organizes points into a ball tree: the leaf level represents the full sequence, and balls in higher levels of the tree represent larger neighborhoods. Erwin enables linear-time attention
by processing nodes in parallel within local neighborhoods of fixed size. It combines fine-grained local attention with progressive pooling for capturing global interactions. This approach performs well at local interactions but may require several steps for distant ones, as accumulating global information requires multiple layers. The progressive coarsening of such hierarchical methods also results in a loss of fidelity, as coarsened features cannot be processed at finer scales.

To address these issues, we introduce Ball Sparse Attention (BSA), which incorporates Ball Tree Attention within NSA's framework to achieve a global receptive field at a sub‐quadratic cost without sacrificing accuracy.

Our contributions are:
\begin{enumerate}[leftmargin=20pt, topsep=-1pt, itemsep=-1pt]
    \item A hybrid architecture integrating Ball Tree Attention into NSA's framework for scalable scientific modeling.
    \item A locality-based sparsification strategy for attention that preserves modeling capacity while reducing computation compared to Full Attention.
    \item A comprehensive validation of Ball Sparse Attention on airflow pressure modeling and stress field prediction in hyperelastic materials. 
\end{enumerate}

%% file: sections/2_methodology.tex
\section{Methodology}
\subsection{Background}
The \textbf{self-attention mechanism} builds on the scaled dot-product attention~\cite{vaswani2017attention}. For an input matrix $X \in \mathbb{R}^{N \times C}$, $X$ is projected into queries, keys, and values:
\begin{align}
Q &= X W_q,\quad K = X W_k,\quad V = X W_v,
\end{align}
where $W_q,\;W_k\;W_v \in \mathbb{R}^{C \times d_k}$ are learnable weight matrices. Then, the attention output is computed as:
\begin{equation}
\mathrm{Attn}(Q, K, V) 
= \mathrm{softmax}\!\Bigl(\tfrac{Q K^T}{\sqrt{d_k}} + \mathcal{B}\Bigr)\,V,
\end{equation}
with $\mathcal{B} \in \mathbb{R}^{N \times N}$ being the bias term. Even with highly optimized implementations (e.g.,~\citealp{dao2022flashattention}), self-attention scales quadratically with the sequence length, making it hard to process sequences longer than tens of thousands of tokens. Recent studies have attempted to address this by imposing geometric or sparsity-based relaxations.

\textbf{Ball Tree Attention:} \citet{zhdanov2025erwintreebasedhierarchicaltransformer} partition the input sequence into disjoint balls $B$ of size $m$, each having $m$ feature vectors $X_B \in \mathbb{R}^{m \times C}$. Then, Ball Tree Attention (BTA) applies scaled-dot-product attention \textit{within} each ball:
\begin{equation}
    X'_B = \mathrm{Attn}^{\mathrm{ball}}(X_B) := Attn\left(X_BW_q, X_BW_k, X_BW_v\right),
\end{equation}
where the weights are shared across balls and maintain row correspondence with $X_B$.

\textbf{Native Sparse Attention (NSA):} 
\citet{yuan2025nativesparseattentionhardwarealigned} preserves full sequence resolution by sparsifying attention in three branches: (1) \textit{compression}, (2) \textit{selection}, and (3) \textit{sliding window}. These branches combine via gated attention:
\begin{equation}
  \mathrm{Attn} = \sum_{b \in \{\mathrm{sld}, \mathrm{cmp}, \mathrm{slc}\}} \sigma(\gamma_b) \odot Attn^b
\end{equation}
where each gate \(\gamma_b\) is passed through a sigmoid function $\sigma(\cdot)$ and used to modulate its branch output $\mathrm{Attn}^b$.

\textit{Compression:} NSA splits $K$ and $V$ into non‐overlapping blocks of length $\ell$ ($\text{stride}=\ell$) and maps each block to a single coarse token via an MLP $\phi$ (or via mean pooling):
\begin{equation}
\begin{aligned}
  K^{\mathrm{cmp}}
    &= \bigl\{ \phi\bigl(K_{(i-1)\ell : i\ell - 1}\bigr)\bigr\}_{i = 1}^{\lceil N / \ell \rceil}, \\
  V^{\mathrm{cmp}}
    &= \bigl\{ \phi\bigl(V_{(i-1)\ell: i\ell - 1}\bigr)\bigr\}_{i = 1}^{\lceil N / \ell \rceil},
\end{aligned}
\end{equation}
where $K_{(i-1)\ell:i\ell - 1}, V_{(i-1)\ell:i\ell - 1}\in\mathbb{R}^{\ell\times d_k}, \forall i\leq \lceil N / \ell \rceil$ \mbox{(0-padded)}, and \(\phi(\cdot)\) outputs a vector in \(\mathbb{R}^{d_k}\). The \emph{compressed attention} is $\mathrm{Attn}^{\mathrm{cmp}} = \mathrm{Attn}(Q, K^{\mathrm{cmp}}, V^{\mathrm{cmp}})$.

\textit{Selection:} For each query position $t$, we reuse the coarse keys to build the dot‐product similarity (importance) matrix:
\begin{equation}
    S = Q (K^{\mathrm{cmp}})^T \in \mathbb R^{N\times\lceil N/\ell\rceil}, \quad S_{tj} = \langle q_t, k^{\mathrm{cmp}}_j\rangle
\end{equation}
The indices of the top $k^\star$ blocks are then selected for each $t$:
\begin{equation}
    \mathcal{I}_t = \operatorname{top-k}(S_{t,.}, k^\star) \subset \{1, \cdots, \lceil N / \ell \rceil\}
\end{equation}
The selected blocks are then converted back
to token-level representations by concatenating their respective KV tokens:
\begin{equation}
\begin{aligned}
    K^{\mathrm{slc}}_t &= \operatorname{Cat}\Bigl\{K_{(i-1)\ell:i\ell - 1} \mid i \in \mathcal{I}_t \Bigr\} \\ 
    V^{\mathrm{slc}}_t &= \operatorname{Cat}\Bigl\{V_{(i-1)\ell:i\ell - 1} \mid i \in \mathcal{I}_t \Bigr\}
\end{aligned}
\end{equation}
The \emph{selection attention} is $\mathrm{Attn}^{\mathrm{slc}}=\mathrm{Attn}(Q, K^{\mathrm{slc}}, V^{\mathrm{slc}})$.

Despite attending to only a fraction of possible pairs, NSA matches or outperforms Full Attention on NLP tasks and achieves an 11$\times$ speedup in computation~\cite{yuan2025nativesparseattentionhardwarealigned}.

\subsection{Ball Sparse Attention}

\textbf{Ball Sparse Attention (BSA):} We propose BSA, an attention mechanism that inherits NSA’s three-branch design: \textit{compression}, \textit{selection}, and \textit{local attention}. However, we replace the conventional sliding window with Ball Tree Attention (BTA)~\cite{zhdanov2025erwintreebasedhierarchicaltransformer} (see BTA in Figure~\ref{fig:figure1} when $q_p = q_t$), allowing the model to attend within continuous geometric regions in $\mathbb{R}^D$ and avoid artificial discontinuities. The compression and selection branches remain as in NSA, and we combine all three branches:
\begin{equation}
  \mathrm{Attn} = \sum_{b \in \{\mathrm{ball}, \mathrm{cmp}, \mathrm{slc}\}} \sigma(\gamma_b) \odot Attn^b.
\end{equation}
\textbf{Group selection:} Beyond locality in BTA, we also exploit locality during \textit{selection}, in the top-$k$ computation. To achieve this, we group query positions $t$ into contiguous groups of size $g$ (see Groping in Figure~\ref{fig:figure1}), and enforce one set of selected blocks across all queries in the same group:
\begin{equation}
    G_p = \bigl\{(p-1)g, \ldots, pg-1\bigr\}, \quad p = 1, \ldots, \bigl\lceil N / g \bigr\rceil,
\end{equation}
where $p$ indexes each group and $G_p$ is the set of indices in the $p$-th group. We also pool the queries in each group:
\begin{equation}
    q_{p} = \frac{1}{|G_p|} \sum_{t \in G_p} Q_t
\end{equation}
We then average the similarity scores within a group and perform top-$k$ selection on these averages:
\begin{equation}
\begin{aligned}
    \bar{S}_{pj} &= \frac{1}{|G_p|} \sum_{t \in G_p} S_{tj}, \quad j = 1, \ldots, \bigl\lceil N / \ell \bigr\rceil,\\
    \mathcal{I}_p &= \operatorname{top\text{-}k}\bigl(\bar{S}_{p,\cdot},\,k^{\star}\bigr), \quad
    \mathcal{I}_t \equiv \mathcal{I}_p \;\; \forall\,t \in G_p.
\end{aligned}
\end{equation}
This reduces top-$k$ calls by a factor of $g$ and allows KV blocks to be fetched in contiguous chunks, improving GPU cache utilisation and lowering memory-access latency.

To further optimize the runtime, we shrink the dimensionality of the query before the \textit{selection} stage by coarsening $Q$ with an operation $\phi$ before computing the similarity matrix. $\phi$ can either be mean pooling or an MLP:
\begin{equation}
    Q^{\mathrm{cmp}} = \bigl\{\;\phi\bigl(Q_{(i-1)\ell : i\ell - 1}\bigr)\bigr\}_{i = 1}^{\lceil N / \ell \rceil},
\end{equation}
with $Q_{(i-1)\ell:i\ell - 1}, \forall i\leq \lceil N / \ell \rceil$ (zero-padded). The smaller similarity matrix and its selection step are then:
\begin{equation}
    \begin{aligned}
    \widetilde{S} &= Q^{\mathrm{cmp}} (K^{\mathrm{cmp}})^T \in \mathbb{R}^{\lceil N / \ell \rceil \times \lceil N / \ell \rceil}, \\ 
    \widetilde{\mathcal{I}}_p &= \operatorname{top-k}(\widetilde{S}_{p,\cdot},k^{\star}) \subset \{1, \cdots, \lceil N / \ell \rceil\},
    \end{aligned}
\end{equation}
after which we proceed exactly as before. This yields a small loss in resolution but improves similarity computation.

\textbf{Group compression:} Finally, we down-sample queries not only for \textit{selection} but also for \textit{compression} itself:
\begin{equation}
     \mathrm{Attn}^{\mathrm{cmp}} = \underbrace{(I_{\lceil N / \ell \rceil} \otimes \mathbf{1}_{\ell})}_{\text{repeat}}\mathrm{Attn}\Bigl(Q^{\mathrm{cmp}}, K^{\mathrm{cmp}}, V^{\mathrm{cmp}}\Bigr),
\end{equation}
where $I_{\lceil N/\ell\rceil}\otimes\mathbf{1}_\ell$ repeats each attention output $\ell$ times. By compressing the queries in both \textit{selection} and \textit{compression}, this variant offers the greatest speed-up at the price of a coarser attention map and a drop in downstream accuracy.

%% file: sections/3_results.tex
\section{Experiments and Results}
\subsection{Experiments setup}
\paragraph{Task} We evaluate the proposed architecture on an airflow pressure modeling task that requires predicting pressure values at discrete points on three-dimensional geometric structures. We use the ShapeNet-Car dataset \cite{shapenet} with 889 car models, each modeled with 3586 surface points in 3D space. The ground truth airflow pressure is generated through simulations with the Reynolds number = $5 \times 10^6$. The data is split into 700 training and 189 testing samples. To test the generalization of our method, we further evaluate with the stress field prediction task on Elasticity benchmark \cite{li2021fourierneuraloperatorparametric}.

\paragraph{Training details} The model consists of 18 transformer blocks, each containing an RMSNorm layer \cite{zhang-sennrich-neurips19}, a Ball Sparse Attention layer (BSA), and a SwiGLU \cite{SWiGLU} for non-linear transformation. The
BSA parameters and training hyperparameters are detailed in Appendix \ref{appendix:training_parameters}. For the query coarsening operation, mean pooling was used for regular BSA, and MLP was used when group compression is used with BSA. To measure FLOPs, we use Deepspeed FLOPs profiler \footnote{\url{https://github.com/deepspeedai/DeepSpeed}}.

\subsection{Experimental results}
\input{figures/receptive_field_visualization}

\begin{minipage}{0.23\textwidth}

    \centering
    \captionof{table}{ShapeNet MSE test performance compared with previous methods.}
    \label{tab:compare_previous_shapenet}
    \vskip 0.15in
    \begin{tabular}{@{}lc@{}}
        \toprule
        \textbf{Model} & \textbf{MSE} \\ \midrule
        PointNet (\citeyear{qi2016pointnet})            & 43.36 \\
        GINO (\citeyear{gino})                & 35.24 \\
        UPT (\citeyear{alkin2024upt})                & 31.66 \\
        Transolver (\citeyear{wu2024Transolver})         & 19.88 \\
        PTv3 (\citeyear{wu2024ptv3})    & 19.09 \\
        GP-UPT (\citeyear{gp-upt})            & 17.02 \\
        Erwin (\citeyear{zhdanov2025erwintreebasedhierarchicaltransformer}) & 15.85 \\ \midrule
        BSA (Ours) & 14.31 \\
        Full Attention (\citeyear{vaswani2017attention})  & 13.29 \\
        \bottomrule
    \end{tabular}
\end{minipage}
\hfill
\begin{minipage}{0.23\textwidth}
\captionof{table}{Elasticity RMSE compared to previous works. All results are multiplied by $10^2$.}
    \label{tab:compare_previous_elasticity}
    \vskip 0.15in
    \begin{tabular}{@{}lc@{}}
        \toprule
        \textbf{Model} & \textbf{RMSE} \\ \midrule
        LSM (\citeyear{wu2023LSM})              &  2.18    \\
        LNO (\citeyear{wang2024LNO})                         & 0.69     \\
        Oformer (\citeyear{oformer})                   & 1.83 \\
        Gnot (\citeyear{hao2023gnot})              & 0.86 \\
        Ono (\citeyear{ono})           & 1.18 \\
        Transolver (\citeyear{wu2024Transolver})   & 0.64 \\
        Erwin (\citeyear{zhdanov2025erwintreebasedhierarchicaltransformer}) & 0.34 \\ \midrule
        BSA (Ours) & 0.38 \\
        Full Attention (\citeyear{vaswani2017attention})  & 0.30 \\
        \bottomrule
    \end{tabular}
\end{minipage}

\paragraph{BSA Comparison with previous methods:}
Table \ref{tab:compare_previous_shapenet} shows that BSA outperforms all previous methods, and has 1.02 MSE worse than Full Attention on ShapeNet. On the stress field prediction task, Table \ref{tab:compare_previous_elasticity} shows that BSA achieves approximately the same performance as Erwin \citep{zhdanov2025erwintreebasedhierarchicaltransformer}. This outcome can be attributed to the small-scale nature of the Elasticity dataset, with sequence lengths of 972, where BSA fails to demonstrate a clear advantage.

\begin{table}[h]
\centering
\caption{Comparison of the Full Attention, BSA, and Erwin in terms of MSE, runtime (ms), and FLOPS on ShapeNet.}
\vskip 0.15in
\label{tab:attention_compared}
\resizebox{\columnwidth}{!}{
  \begin{tabular}{lccc}
    \toprule
    \textbf{Attention type} & \textbf{MSE}   & \textbf{Runtime (ms)} & \textbf{GFLOPS} \\ 
    \midrule
    Erwin 
      & 16.12 & 19.35        & 14.60   \\ 
    Full Attention 
      & 13.29 & 37.82        & 87.08  \\ \midrule
    BSA & 14.31 & 36.53 & 27.91 \\
    BSA w/o group selection  & 14.44 & 66.92        & 32.67  \\
    BSA w group compression & 14.80 &  23.42 & 20.82 \\
    
    \bottomrule
  \end{tabular}
}
\end{table}

\paragraph{Accuracy-Efficiency trade-off:} Table~\ref{tab:attention_compared} shows that all BSA variants achieve significantly better results than Erwin~\cite{zhdanov2025erwintreebasedhierarchicaltransformer} and approach the accuracy of Full Attention~\cite{vaswani2017attention}. Although these variants have higher GFLOPs than Erwin, they require fewer GFLOPs than Full Attention. However, when measuring the runtime on ShapeNet with a sequence length of 4096, Full Attention outperforms BSA without group selection. This is because, unlike NSA \cite{yuan2025nativesparseattentionhardwarealigned}, we do not implement a Triton kernel for efficient selection. 

Table~\ref{tab:attention_compared} also illustrates the trade-off between accuracy and computational efficiency across different BSA variants. While BSA without group selection has higher GFLOPS and runtime, the difference in MSE is marginal (only 0.13). This supports our assumption about the importance of local structures in physical tasks, showing that efficiency can be improved without a significant change in accuracy. Furthermore, a preliminary analysis of how the compressed block size $\ell$ and the group selection size $|G_p|$ influence ShapeNet performance is provided in Appendix \ref{appendix:block-ablations}.

\input{figures/bsa_vs_full_attn_scaling}
\paragraph{BSA scaling} Figure~\ref{fig:sparse_full_scaling} shows the runtime
scaling of our BSA with sequence length compared to
Full Attention. At smaller sequence lengths, full attention
exhibits faster runtime due to the computational overhead at
the MLP of BSA. However, as the sequence length increases,
the gap narrows, with BSA outperforming Full Attention at
4096 sequence length. At the longest sequence length of
65536, BSA is 5× faster than full attention. Appendix \ref{appendix:scaling_analysis} provides a detailed scaling comparison of the BSA variants. This highlights
the superior computational efficiency of our BSA at scale.

\paragraph{BSA receptive field} Figure \ref{fig:receptive_field} demonstrates the receptive field of each individual component in BSA. Initially, Ball Tree Attention (BTA) only attends to points inside a local ball. With selection, it can attend to different blocks far away from the current point. Additionally, to prevent the model from over‐fitting to local patterns already captured by the BTA, we mask all blocks within each ball, encouraging it to attend to more distant regions. Finally, with compression, it achieves a global receptive field by attending to all coarse key–value representations of each local block.

%% file: figures/receptive_field_visualization.tex
\begin{figure}[]
  \centering
  \renewcommand{\arraystretch}{1.1} 

  \begin{tabular}{ccc}
    \includegraphics[width=.15\textwidth]{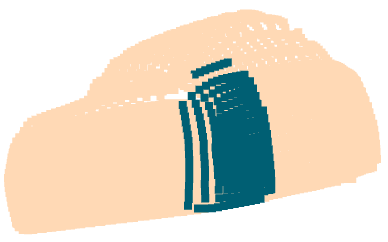} &
    \includegraphics[width=.15\textwidth]{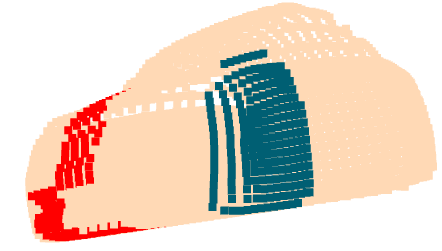} &
    \includegraphics[width=.15\textwidth]{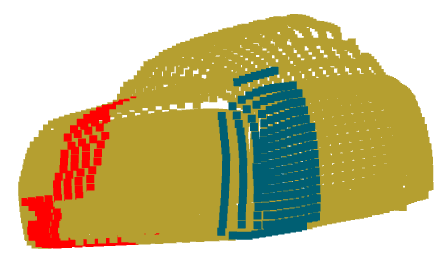} \\[4pt]

    Ball attention &
    + Selection &
    + Compression \\
  \end{tabular}

  \caption{Receptive field visualization of a car in the Shapenet dataset with different components: ball attention; ball attention and selection; ball attention, selection and compression. The receptive field increases with more components.}
  \label{fig:receptive_field}
\end{figure}

%% file: figures/bsa_vs_full_attn_scaling.tex
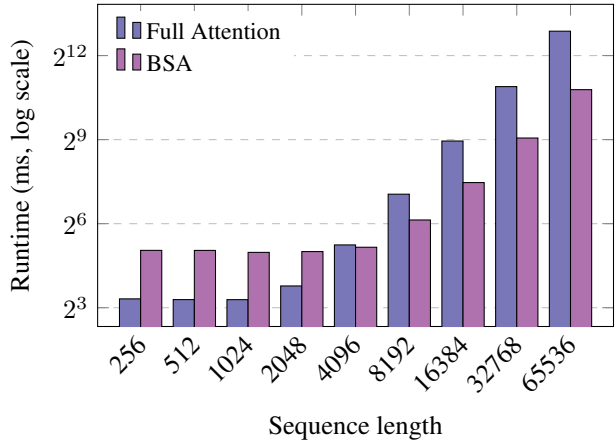
\begin{figure}[]
\centering
\begin{tikzpicture}
\begin{axis}[
    width=0.4\textwidth,
    height=0.25\textwidth,
    xmode=log,
    log basis x=2,
    ymode=log,
    log basis y=2,
    xlabel={Sequence length},
    ylabel={Runtime (ms, log scale)},
    legend style={draw=none, font=\small, anchor=north west, at={(0.02,0.98)}},
    legend cell align=left,
    xtick={256,512,1024,2048,4096,8192,16384,32768,65536},
    xticklabels={256,512,1024,2048,4096,8192,16384,32768,65536},
    xticklabel style={rotate=45, anchor=east, yshift=-0.4em},
    ymajorgrids=true,
    grid style=dashed,
    scale only axis,
]

\addlegendimage{ybar, ybar legend, fill=Periwinkle, draw=black}
\addlegendentry{Full Attention}
\addlegendimage{ybar, ybar legend, fill=Orchid, draw=black}
\addlegendentry{BSA}
\addplot[ybar, fill=Periwinkle, bar width=8pt, bar shift=-4pt] plot coordinates {
    (256,9.958009143) (512,9.778074083) (1024,9.768114715) (2048,13.70764882) (4096,37.82152729) (8192,132.7062093) (16384,494.3802367) (32768,1899.925044) (65536,7484.721494)
};
\addplot[ybar, fill=Orchid, bar width=8pt, bar shift=4pt] plot coordinates {
    (256,33.061312465667726) (512,33.03422769546509) (1024,31.454048843383788) (2048,32.08338500022888) (4096,35.70635498046875) (8192,70.12005973815918) (16384,176.85500732421875) (32768,532.5928186035156) (65536,1761.63009765625)

};

\end{axis}

\end{tikzpicture}
\vspace{-0.5em}
\caption{Runtime of BSA and Full attention with increasing sequence length (from 256 to 65536).}
\label{fig:sparse_full_scaling}
\end{figure}

%% file: sections/4_conclusion.tex
\section{Conclusion}
We introduce BSA, a novel sparse attention mechanism suited for large scale physical systems. Our approach integrates Ball Tree Attention within the NSA framework and introduces a locality-based sparsification strategy that preserves the performance of high-quality long-range modeling while drastically reducing computational efficiency. Through grouped selection and hybrid attention, we preserve local and global context with minimal overhead.

Our results show that BSA achieves accuracy competitive with both Full Attention and the Erwin Transformer~\cite{zhdanov2025erwintreebasedhierarchicaltransformer}. Notably, at larger sequence lengths, BSA is up to five times faster than Full Attention. Its hybrid attention mechanism expands the receptive field compared to Erwin, enabling better long-range information propagation. Given its efficiency and performance at scale, BSA is ideal for simulating large-scale physical systems.

\paragraph{Future work:} In the future, we will evaluate our fixed‐group query partitioning scheme on a broad spectrum of point‐cloud datasets to assess its robustness across domains, conduct comprehensive hyperparameter sweeps and ablation studies to quantify BSA’s sensitivity to tuning. Moreover, we plan to develop a GPU kernel for BSA operations to fully make use of NSA's hardware alignment for improved computational efficiency and reduced memory footprint.

%% file: sections/5_appendix.tex
\onecolumn
\section{Training hyperparameters}
\label{appendix:training_parameters}
Table \ref{tab:sparse_params} details the sparse attention hyperparameters. The model is trained with mean squared error loss for the regression task, and it is trained for 100000 iterations with AdamW optimizer \cite{loshchilov2019decoupledweightdecayregularization} with a cosine learning rate scheduler, learning rate 0.001, and weight decay 0.01.
\begin{table}[h!]
\centering
\caption{Sparse attention parameters}
\label{tab:sparse_params}
\vskip 0.15in
\begin{tabular}{ll}
\hline
Parameter & Value \\ \hline
Ball size & 256 \\
Compression block size & 8 \\
Compression block sliding stride & 8 \\
Selection block size & 8 \\
Number of blocks selected & 4 \\ \hline
\end{tabular}
\end{table}

\section {Block size ablations on ShapeNet}
\label{appendix:block-ablations}
\begin{table}[ht]
\centering
\caption{BSA test MSE for various compression and group selection sizes on ShapeNet. We use $k$=4 (for top-$k$) with mean pooling.}
\label{tab:blocksize_mse}
\vskip 0.15in
\begin{tabular}{ccc}
\toprule
Compr. Block Size & Group Selection Size & Val. MSE \\
\midrule
4  & 4  &  15.43  \\
8  & 8  &  14.31  \\
16 & 16 &  14.97  \\
32 & 32 & 132.14  \\
\midrule
4  & 8  &  14.81  \\
16 & 8  &  14.88  \\
8  & 4  &  14.88  \\
8  & 16 &  14.84  \\
\bottomrule
\end{tabular}
\end{table}

\section{Runtime scaling analysis}
\label{appendix:scaling_analysis}

\input{figures/bsa_variants_scaling}

%% file: figures/bsa_variants_scaling.tex
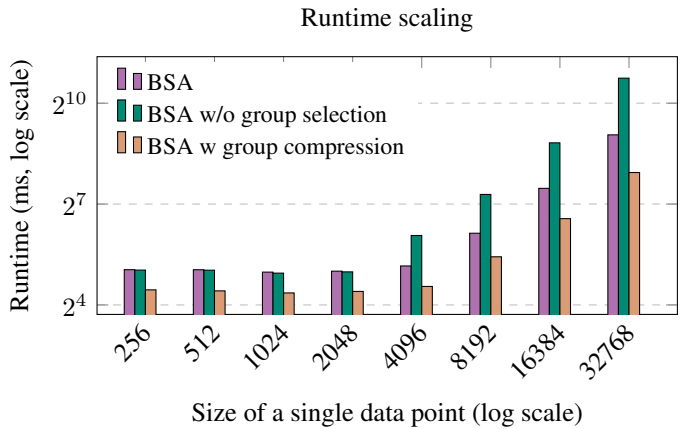
\begin{figure}[ht]
\centering
\begin{tikzpicture}
\begin{axis}[
    width=0.45\textwidth,
    height=0.20\textwidth,
    xmode=log,
    log basis x=2,
    ymode=log,
    log basis y=2,
    xlabel={Size of a single data point (log scale)},
    ylabel={Runtime (ms, log scale)},
    title={Runtime scaling},
    legend style={draw=none, font=\small, anchor=north west, at={(0.02,0.98)}},
    legend cell align=left,
    xtick={256,512,1024,2048,4096,8192,16384,32768},
    xticklabels={256,512,1024,2048,4096,8192,16384,32768},
    xticklabel style={rotate=45, anchor=east, yshift=-0.4em},
    ymajorgrids=true,
    grid style=dashed,
    scale only axis
]

\addlegendimage{ybar, ybar legend, fill=Orchid, draw=black}
\addlegendentry{BSA}
\addlegendimage{ybar, ybar legend, fill=PineGreen, draw=black}
\addlegendentry{BSA w/o group selection}
\addlegendimage{ybar, ybar legend, fill=Tan, draw=black}
\addlegendentry{BSA w group compression}

\addplot[ybar, fill=Orchid, bar width=4pt, bar shift=-6pt] plot coordinates {
    (256,33.061312465667726) (512,33.03422769546509) (1024,31.454048843383788) (2048,32.08338500022888) (4096,35.70635498046875) (8192,70.12005973815918) (16384,176.85500732421875) (32768,532.5928186035156)
};
\addplot[ybar, fill=PineGreen, bar width=4pt, bar shift=-2pt] plot coordinates {
    (256,32.7986427116394) (512,32.73433200836182) (1024,30.808453855514525) (2048,31.611333856582643) (4096,66.92052406311035) (8192,155.98654235839842) (16384,451.7363325500488) (32768,1713.304896850586)
};
\addplot[ybar, fill=Tan, bar width=4pt, bar shift=2pt] plot coordinates {
    (256,21.824155817031862) (512,21.370238275527953) (1024,20.481154747009278) (2048,21.132762670516968) (4096,23.417129821777344) (8192,43.14087310791015) (16384,94.66729671478271) (32768,245.07135284423828)
};

\end{axis}
\end{tikzpicture}
\vspace{-0.5em}
\caption{Runtime of BSA and its variants with increasing sequence length (from 256 to 32768).}
\end{figure}